\documentclass{article}
\usepackage{ijcai26}
\usepackage{times}
\usepackage{soul}
\usepackage[breaklinks]{hyperref}
\usepackage{url}
\usepackage[utf8]{inputenc}
\usepackage[small]{caption}
\usepackage{graphicx}
\usepackage{pdfpages}
\usepackage{amsmath}
\usepackage{amsthm}
\usepackage{booktabs}
\usepackage{algorithm}
\usepackage{algorithmic}
\usepackage[switch]{lineno}
\title{BlenderRAG: High-Fidelity 3D Object Generation\\via Retrieval-Augmented Code Synthesis}

\author{
Massimo Rondelli$^{1}$
\and
Francesco Pivi$^{1,2}$
\and
Maurizio Gabbrielli$^{1}$
\affiliations
$^1$University of Bologna, Bologna, Italy\\
$^2$Ferrari S.p.A., Maranello, Italy\\
\emails
\{massimo.rondelli2,
francesco.pivi2\}@unibo.it
}

\begin{document}

\maketitle

\begin{abstract}
Automatic generation of executable Blender code from natural language remains challenging, with state-of-the-art LLMs producing frequent syntactic errors and geometrically inconsistent objects. We present BlenderRAG, a retrieval-augmented generation system that operates on a curated multimodal dataset of 500 expert-validated examples (text, code, image) across 50 object categories. By retrieving semantically similar examples during generation, BlenderRAG improves compilation success rates from $40.8\%$ to $70.0\%$ and semantic normalized alignment from $0.41$ to $0.77$ (CLIP similarity) across four state-of-the-art LLMs, without requiring fine-tuning or specialized hardware, making it immediately accessible for deployment. The dataset and code will be available at \url{https://github.com/MaxRondelli/BlenderRAG}.
\end{abstract}

\section{Introduction}

Generating executable Blender Python code from textual descriptions poses significant challenges for Large Language Models ~\cite{chen2021evaluating,austin2021program,roziere2023code,li2023starcoder}. Current approaches suffer from syntactic errors, inconsistent proportions, and poor geometric coherence. BlenderLLM \cite{du2024blenderllm} addresses this through iterative fine-tuning with self-improvement, but requires expensive GPU resources and complex training pipelines. SceneCraft \cite{yang2024scenecraft} uses multi-agent decomposition for scene composition but prioritizes structure over visual quality of individual objects, while 3D-GPT \cite{sun20253d} focuses on procedural modeling through LLM-based planning.
\begin{figure}[t]
    \centering
    \includegraphics[width=\linewidth]{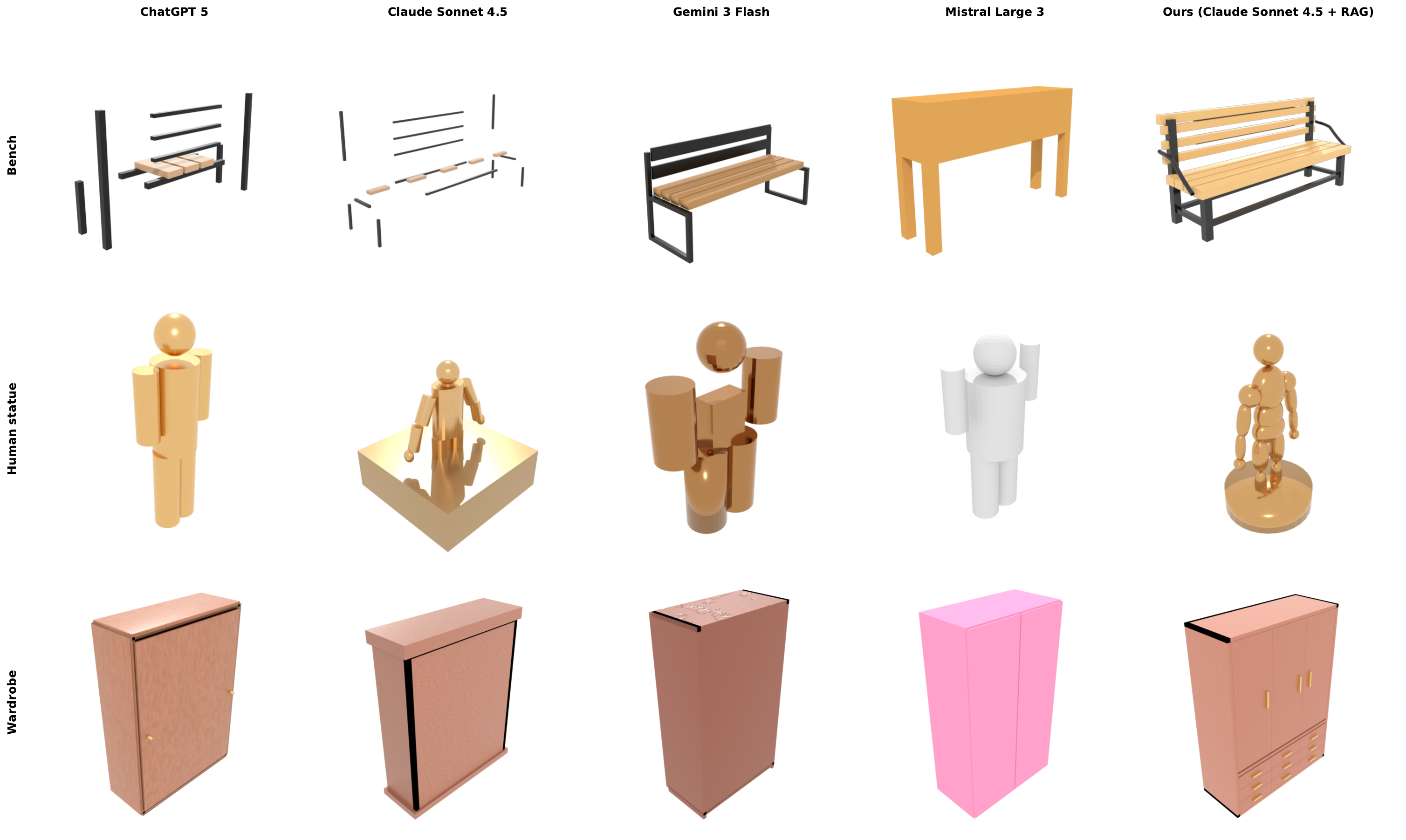}
    \caption{Qualitative comparison: BlenderRAG with as backbone model Claude Sonnet 4.5 vs baseline models. RAG outputs exhibit superior geometric coherence, realistic proportions, and structural detail.}
    \label{fig:comparison}
\end{figure}
BlenderRAG offers a complementary approach: a deployment-ready system requiring no fine-tuning that prioritizes geometric coherence and visual realism through retrieval-augmented generation \cite{lewis2020retrieval,guu2020retrieval,borgeaud2022improving}. 
Unlike approaches requiring GPU-intensive fine-tuning
or specialized training infrastructure, BlenderRAG operates as a zero-training system deployable on standard consumer hardware. This design choice prioritizes accessibility and immediate usability over methods requiring substantial computational investment for model adaptation.
By providing semantically similar expert-validated examples as context, the system guides diverse LLMs to produce high-quality 3D objects with minimal computational overhead.

\section{Methodology}
\subsection{Dataset}
We constructed a multimodal dataset of 500 examples across 50 object categories (25 indoor, 25 outdoor), each with 10 design variations. Each instance comprises: (1) detailed textual description, (2) executable Blender Python code, (3) rendered 2D image.
\begin{itemize}
    \item \textbf{Indoor Categories (25):} Armchair, Bed, Bookshelf, Cabinet (Kitchen), Candle, Chair, Door, Frame, Fridge, Glass, Lamp, Living Room Table, Microwave, Mirror, Office Lamp, Pillow, Plant, Plate, Pot, Rug, Sofa, Table, Trash Can, Wardrobe, Window.
    \item \textbf{Outdoor Categories (25):} Ball, Bell Tower, Bench, Bin, Bush, Cactus, Car, Condominium, Daisy, Fountain, Gate, Gazebo, Grass, Hedge, Humanoid Statue, Mountain, Rock, Sea Umbrella, Shrub, Shrub Leaf, Skyscraper, Stop Signal, Stoplight, Street Lamp, Tree.
\end{itemize}
\begin{figure*}[t]
    \centering
    \includegraphics[width=0.95\linewidth]{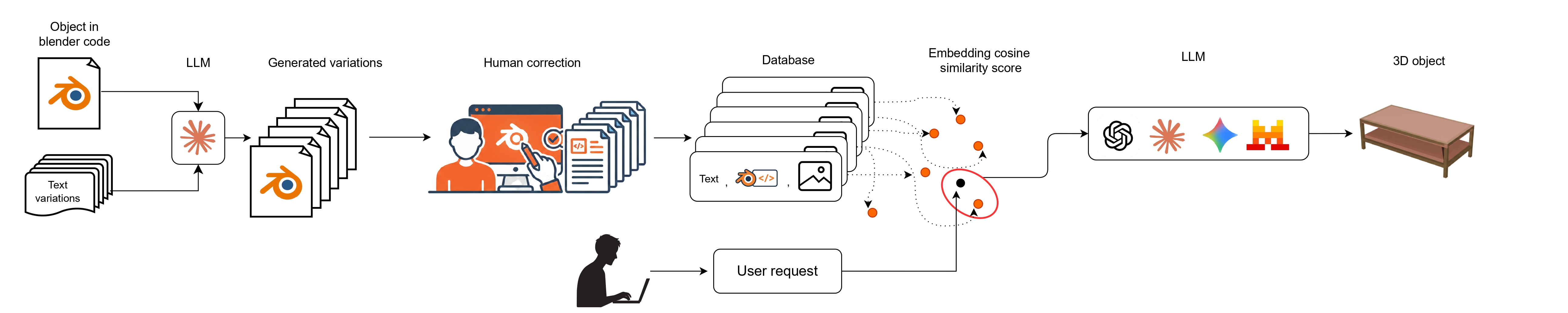}
    \caption{BlenderRAG architecture: user queries (text/image) are embedded and matched against the Qdrant vector database. Retrieved triplets (text-code-image) provide context for LLM generation, producing executable Blender code through the integrated plugin interface.}
    \label{fig:pipeline}
\end{figure*}

\textbf{Generation Pipeline.} Initial code drafts were generated by Claude Opus 4.1 \cite{anthropic2025claude41} from detailed prompts, then manually validated and refined by expert users to ensure geometric accuracy and visual realism. Images were rendered with standardized camera positioning: front-right-top view (45° horizontal, 30° vertical elevation in spherical coordinates), with adaptive zoom ensuring complete object capture. Objects are centered with uniform lighting.

\begin{figure}[h]
    \centering
    \includegraphics[width=\linewidth]{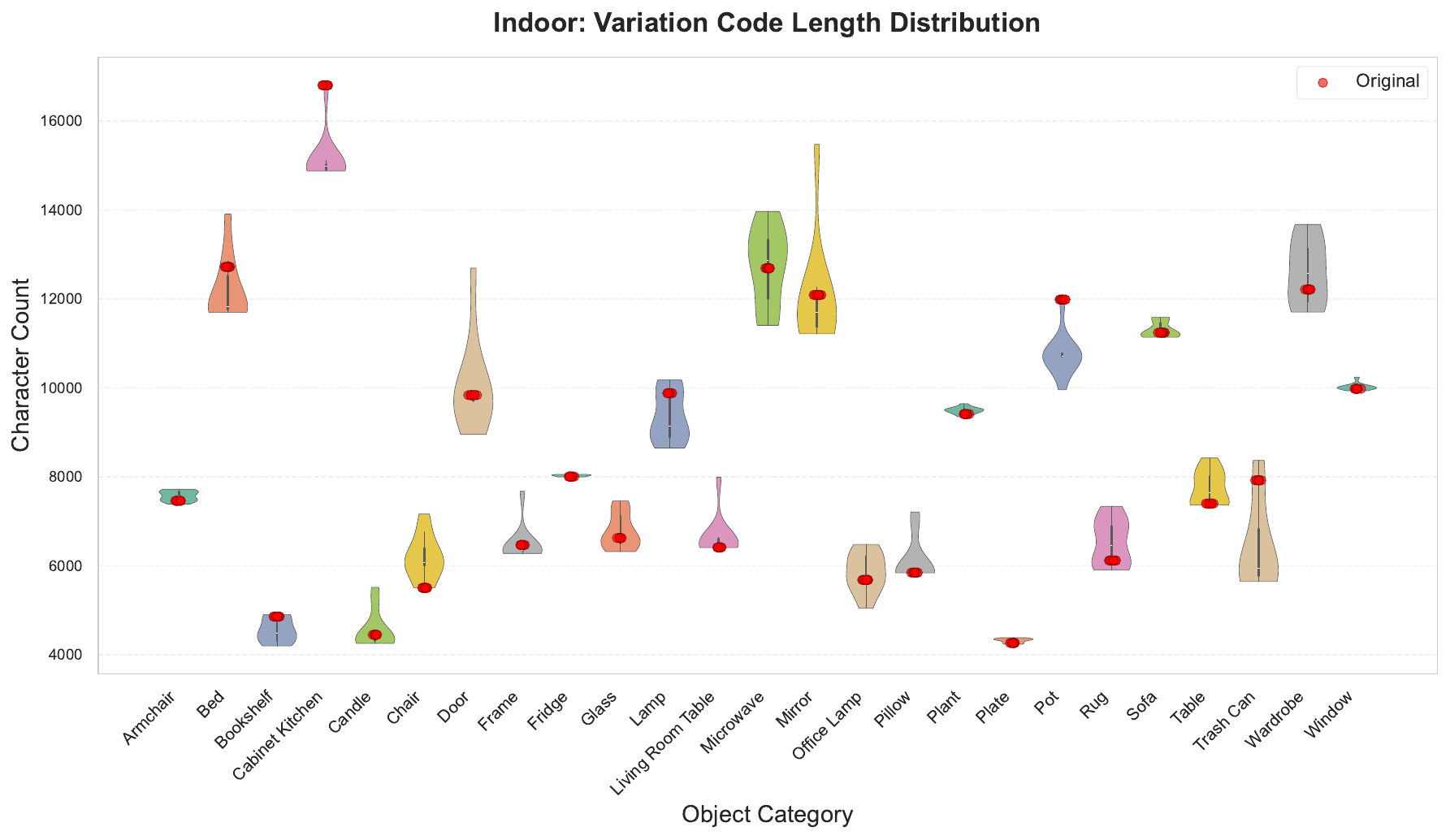}
    \caption{Code length distribution for indoor objects. Red points show original templates. Complexity ranges from simple objects like Plate ($\sim$4,000 chars) to complex structures like Cabinet Kitchen ($\sim$15,000 chars).}
    \label{fig:indoor}
\end{figure}

\begin{figure}[h]
    \centering
    \includegraphics[width=\linewidth]{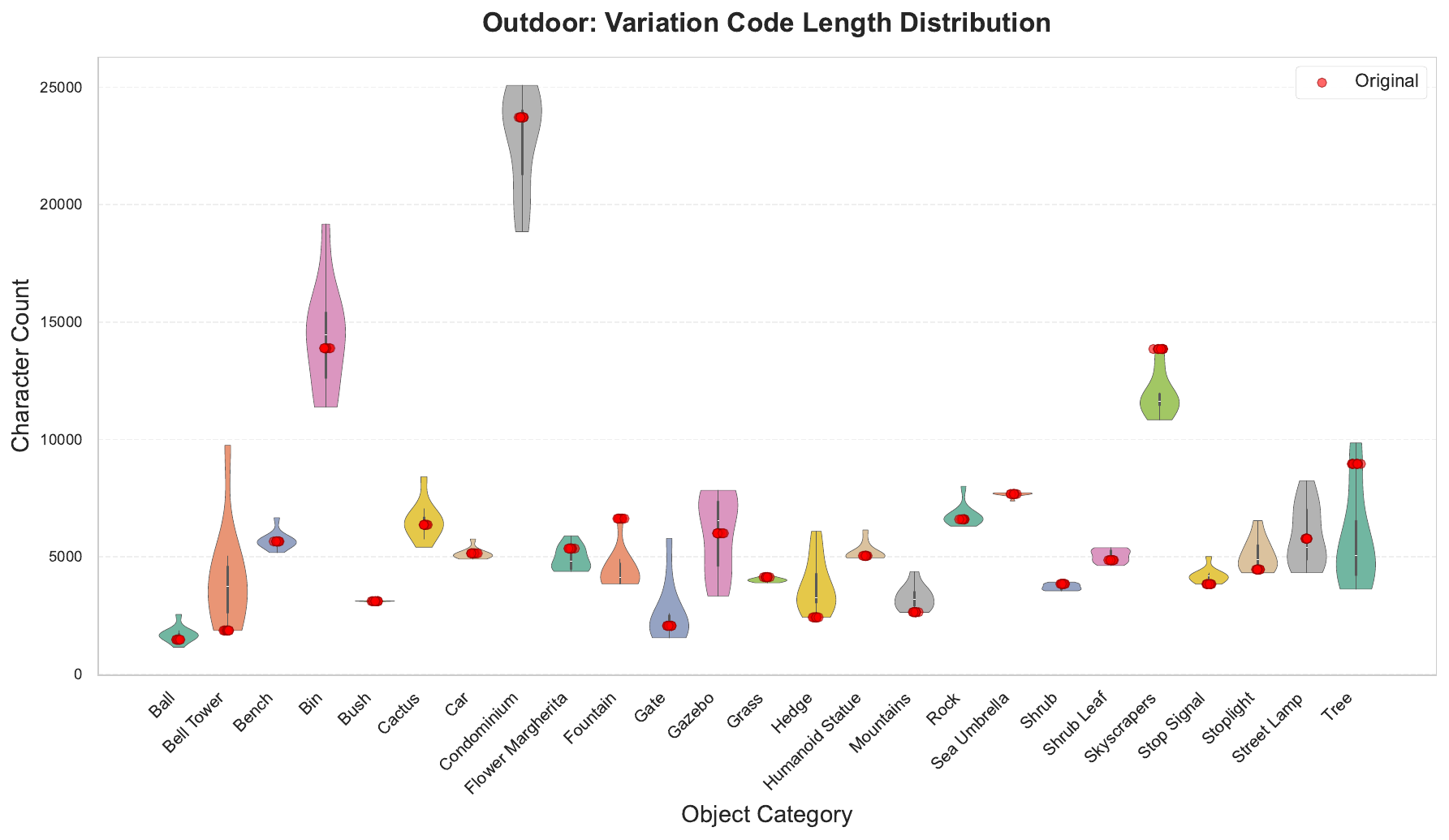}
    \caption{Code length distribution for outdoor objects. Red points show original templates. Notable variation from simple Ball ($\sim$3,000 chars) to elaborate Condominium ($\sim$24,000 chars), capturing diverse structural complexity.}
    \label{fig:outdoor}
\end{figure}

\textbf{Dataset Analysis.} Figures~\ref{fig:indoor} and \ref{fig:outdoor} show code complexity distributions. Indoor objects exhibit moderate complexity (4,000--15,000 characters), while outdoor objects span wider ranges (3,000--24,000 characters) reflecting architectural diversity. This variation ensures the retrieval mechanism must identify appropriate examples across diverse complexity levels.

\subsection{Retrieval-Augmented Generation}

The dataset is indexed in a Qdrant vector database using Nomic-AI embeddings for textual descriptions. During inference, user queries are embedded and matched semantically; the system retrieves the \textit{k} most similar examples. The retrieved text descriptions and object code are then injected into the LLM prompt as context, providing structural code patterns for generation.

The system integrates as a Blender plugin supporting multiple LLM backends (Claude Sonnet 4.5 \cite{anthropic2025claude41}, GPT-5 \cite{openai2025gpt5}, Gemini 3 Flash \cite{google2026gemini3flash}, Mistral Large \cite{mistral2025large3}). Generated code executes directly in Blender's Python environment, enabling immediate 3D object creation.
\subsection{Zero-Training Deployment Philosophy}

BlenderRAG deliberately avoids model fine-tuning to ensure broad accessibility. While fine-tuning approaches like BlenderLLM \cite{du2024blenderllm} achieve strong results, they require: (1) multi-GPU training infrastructure, (2) substantial computational budgets (hundreds of GPU-hours), and (3) expertise in training pipeline configuration. These requirements create barriers for individual artists, small studios, and educational settings.

In contrast, BlenderRAG runs entirely on CPU with any off-the-shelf LLM API, requiring only vector database indexing and minimal memory overhead for embeddings. This enables deployment on consumer laptops without specialized hardware.

\subsection{Blender Add-on Integration}

BlenderRAG is deployed as a native Blender add-on that perfectly integrates into the 3D modeling workflow. The interface allows users to select their preferred LLM client (Claude Sonnet 4.5, GPT-5, Gemini 3 Flash, or Mistral Large) and input textual descriptions of desired objects.

The generation pipeline operates as follows: (1) the user's input is embedded using Nomic-AI; (2) the three most semantically similar examples are retrieved from the Qdrant database; (3) retrieved text descriptions and object code are injected into the LLM prompt as context; (4) the selected model generates executable Python code; (5) the code is automatically executed in the current Blender scene; (6) a 3D rendering of the generated object is produced and displayed.  Users can iteratively refine results through follow-up prompts or directly edit the generated code for fine-grained control.
\begin{figure}[ht]
    \centering
    \includegraphics[width=1\linewidth]{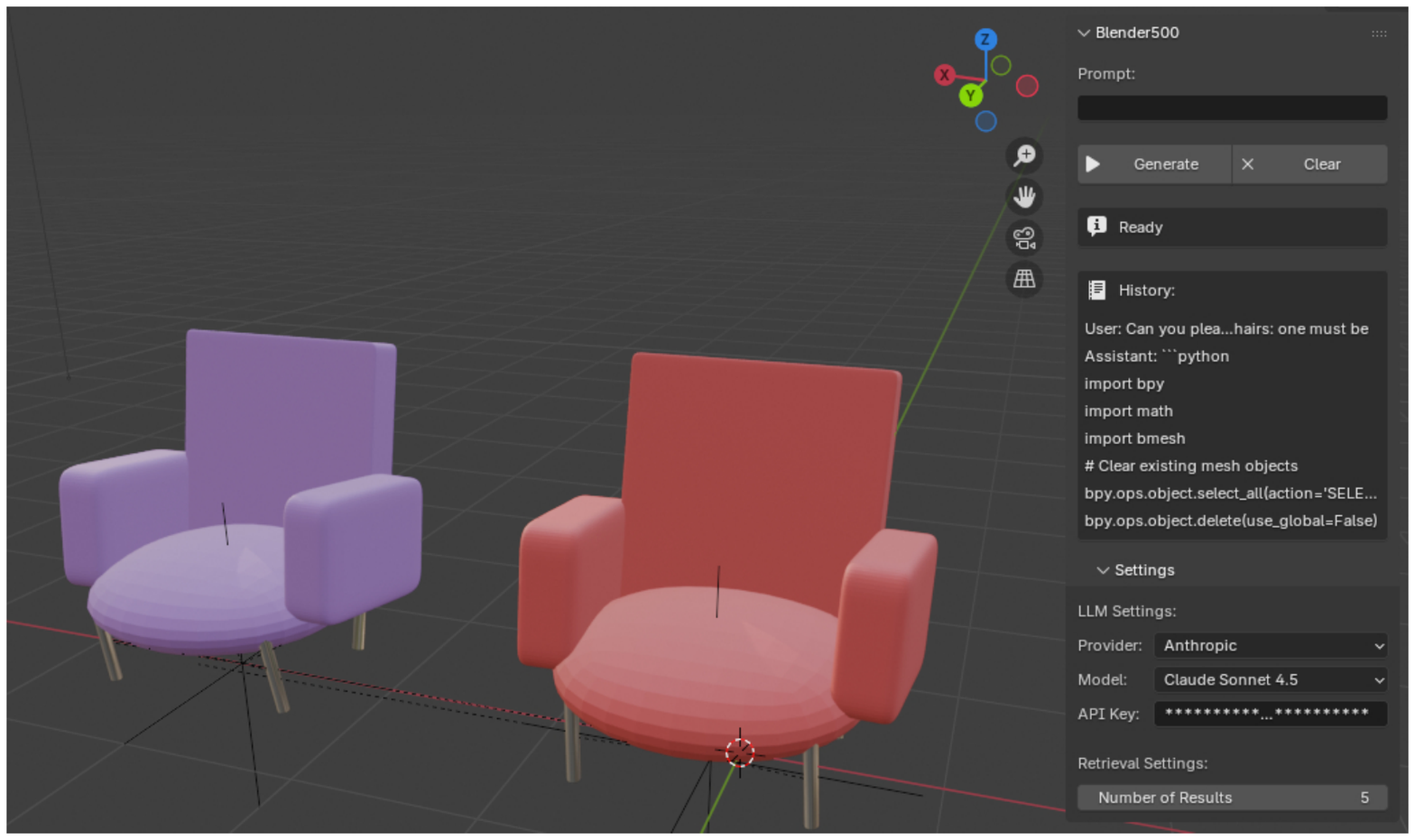}
    \caption{BlenderRAG add-on interface integrated in Blender, showing model selection and prompt input panel. }
    \label{fig:addon}
\end{figure}

\section{Experimental Evaluation}

\subsection{Setup}

We evaluated on 30 out-of-distribution prompts describing novel objects absent from the dataset. Performance measured via: (1) \textbf{Compilation Success Rate}: proportion of generated code executing without errors; (2) \textbf{Semantic Alignment}: CLIP-based normalized cosine similarity between input prompt and rendered output image. 

\begin{table}[h]
\centering
\caption{Compilation success rates (\%) and semantic alignment scores (normalized CLIP cosine similarity) across models. BlenderRAG consistently improves both metrics. Base indicate the simple pure performance in code generation of the selected model while RAG indicates the prompt injection of the correctly retrieved objects.}
\label{tab:results}
\small
\begin{tabular}{lcccc}
\toprule
& \multicolumn{2}{c}{\textbf{Compilation (\%)}} & \multicolumn{2}{c}{\textbf{Alignment}} \\
\cmidrule(lr){2-3} \cmidrule(lr){4-5}
\textbf{Model} & Base & RAG & Base & RAG \\
\midrule
Claude Sonnet 4.5 & $43.3$ & $76.7$ & $0.544$ & $0.780$ \\
GPT-5 Chat & $56.6$ & $66.7$ & $0.267$ & $0.777$ \\
Gemini 3 Flash & $53.3$ & $80.0$ & $0.498$ & $0.770$ \\
Mistral Large & $10.1$ & $56.7$ & $0.327$ & $0.769$ \\
\midrule
\textbf{Average} & $40.8$ & $70.0$ & $0.409$ & $0.774$ \\
\bottomrule
\end{tabular}
\end{table}

\subsection{Results}

Table~\ref{tab:results} shows BlenderRAG substantially improves both compilation rates (from $40.8\%$ to $70.0\%$) and normalized semantic alignment (from $0.409$ to $0.774$ similarity score) across all models. Without retrieval, models achieve 10--56\% compilation success; with RAG, success rates reach $80$\%, approaching practical utility, and being a substantial increase for all four models. Semantic alignment improvements indicate retrieved examples guide not only syntactic correctness but also appropriate geometric and material properties.

Figure~\ref{fig:comparison} demonstrates qualitative improvements: RAG outputs show realistic articulation, proper dimensional relationships, and structurally sound elements, while baselines exhibit floating components, inconsistent scales, and geometric implausibility.

\section{Conclusion}

BlenderRAG demonstrates that retrieval-augmented generation over expert-curated multimodal examples substantially improves 3D object generation quality without fine-tuning. The system achieves practical compilation rates of $80.0\%$ using Gemini 3 Flash and strong normalized semantic alignment CLIP similarity across all LLMs through simple semantic retrieval. Integrated as a Blender plugin, it enables immediate deployment for content creation workflows. 
Future work will include: (1) extending the system to multi-object scene composition with spatial reasoning, (2) incorporating active learning mechanisms for continuous dataset expansion based on user generations, and (3) exploring text-to-image retrieval to enable image-based query input alongside textual descriptions. These enhancements would further improve the system's practical utility while maintaining its zero-training deployment philosophy.

\section*{Acknowledgments}
We thank expert modelers for dataset validation and reviewers for constructive feedback.
\bibliographystyle{named}  %
\bibliography{ijcai26}  %

\end{document}